\documentclass[11pt,a4paper]{article}
\usepackage[hyperref]{acl2020}
\usepackage{times}
\usepackage{latexsym}
\usepackage{graphicx}
\usepackage{booktabs}
\usepackage{pifont}
\usepackage[htt]{hyphenat}
\usepackage{amssymb}
\usepackage{amsmath}
\usepackage{hyperref}
\usepackage{multirow}

\newcommand{\bftab}{\fontseries{b}\selectfont}

\usepackage{microtype}

\aclfinalcopy 

\title{Pseudo-Labels Are All You Need}

\author{Bogdan Kostić \and Mathis Lucka \and Julian Risch \\
deepset\\ \texttt{\{bogdan.kostic, mathis.lucka, julian.risch\}@deepset.ai}}
\date{}

\begin{document}
\maketitle
\begin{abstract}
Automatically estimating the complexity of texts for readers has a variety of applications, such as recommending texts with an appropriate complexity level to language learners or supporting the evaluation of text simplification approaches.
In this paper, we present our submission to the Text Complexity DE Challenge 2022, a regression task where the goal is to predict the complexity of a German sentence for German learners at level B.
Our approach relies on more than 220,000 pseudo-labels created from the German Wikipedia and other corpora to train Transformer-based models, and refrains from any feature engineering or any additional, labeled data.
We find that the pseudo-label-based approach gives impressive results yet requires little to no adjustment to the specific task and therefore could be easily adapted to other domains and tasks.
\end{abstract}



\section{Introduction}
\label{sec:introduction}
What makes some texts more difficult to read for learners of a foreign language than others? How does a complicated sentence construction or the use of rare vocabulary increase complexity? The prediction of text complexity with machine learning methods addresses these questions. 
In contrast to last years' shared tasks at KONVENS, which focused on the disambiguation of German verbal idioms~\cite{ehren-etal-2021-shared}, the identification of toxic, engaging, and fact-claiming comments~\cite{risch-etal-2021-overview}, and scene segmentation in narrative texts~\cite{zehe2021shared}, the task of 2022 is about text complexity.
In this paper, we present our submission to this Text Complexity DE Challenge 2022.
It is a shared task addressing the automatic estimation of the complexity of German sentences for readers, in particular, German learners at level B.
The provided training dataset contains about 1000 sentences and the test dataset about 300 sentences in German.
Figure~\ref{fig:placeholder} shows an exemplary sentence from the shared task dataset in German, an English translation, and the arithmetic mean of ratings from all annotators.
With a seven-level Likert-scale with values ranging from very easy (1) to very complex (7), this task is a regression task and it is evaluated using the Root Mean Squared Error (RMSE).
A third-order mapping is applied before the error is measured so that the impact of any systematic bias in the predictions on the metrics is reduced.
Thereby, the focus of the evaluation is shifted towards ranking sentences correctly with regards to their complexity rather then assigning the correct absolute complexity score.
We refer to the overview paper of the shared task for more details about the dataset and the overall results~\cite{mohtaj-etal-2021-overview}.

\setlength{\fboxsep}{8pt}
\begin{figure}
\centering
\fbox{\parbox{0.90\linewidth}{
\textbf{German Sentence:} 
Als Versauerung der Meere wird die Abnahme des pH-Wertes des Meerwassers bezeichnet.\\
\textbf{English Translation:} 
Ocean acidification is the term used to describe the decrease in the pH of seawater.\\
\textbf{Compexity Score:} 2.13
}}
\fbox{\parbox{0.90\linewidth}{
\textbf{German Sentence:} 
Nach chemischer Härtung des Rußes war er in der Lage, auf galvanoplastischem Wege ein Zink-Positiv und von diesem ein Negativ der Platte anzufertigen, das als Stempel zur Pressung beliebig vieler Positive genutzt werden konnte – die Schallplatte war erfunden.\\
\textbf{English Translation:} 
After chemical hardening of the carbon black, he was able to produce a zinc positive by galvanoplastic means and from this a negative of the record, which could be used as a stamp for pressing any number of positives - the record was invented.\\
\textbf{Complexity Score:} 4.70
}}
  \caption{Two sentences from the training dataset.}\label{fig:placeholder}%
\end{figure}

The remainder of this paper is structured as follows.
Section~\ref{sec:related-work} summarizes related work on text complexity estimation and on pseudo-labeling techniques for machine learning.
We describe our approach in Section~\ref{sec:approach} and its evaluation in Section~\ref{sec:experiments}, with experiments on the validation dataset provided by the shared task organizers.
We conclude in Section~\ref{sec:conclusion} and provide an outlook on future work.

\section{Related Work}
\label{sec:related-work}
Research on reading complexity of German texts is so far relatively scarce with several papers introducing datasets of annotated German sentences or longer texts and mostly feature-based approaches for text complexity prediction.
First of all, there is a dataset with sentence-level annotations, which is the basis of this shared task~\cite{naderi-etal-2019-subjective}.
\citet{rios-etal-2021-new} introduce a dataset for document-level text complexity with the application focus of text simplification and there are two other document-level text complexity datasets by \citet{battisti-etal-2020-corpus} and by \citet{hewett-stede-2021-automatically}. 
The latter follows the format of a similar study~\cite{hulpus-etal-2019-spreading} based on a dataset of English newspaper articles~\cite{10.1162/tacl_a_00139}.
Another dataset is from a Kaggle challenge called CommonLit Readability Prize, where the task is to rate the complexity of literary passages for school grades 3-12.\footnote{\url{https://www.kaggle.com/competitions/commonlitreadabilityprize/}}
Last but not least, there are unlabeled datasets of German texts with simple language, such as the Tagesschau/Logo corpus and the Geo/Geolino corpus~\cite{weiss-meurers-2018-modeling} or Klexikon~\cite{aumiller-gertz-2022-klexikon}.
These datasets cannot be used directly for fine-tuning models on the text complexity prediction task due to the lack of annotations.
However, we show in our approach that they can be used in combination with pseudo-labeling.

Similar to the pseudo-labeling approach that we use, there is a data augmentation technique where a slow but more accurate cross-encoder model is used to label a large set of otherwise unlabeled data samples~\cite{thakur-etal-2021-augmented}. 
This technique augments the training data for a faster, less complex bi-encoder model to address a pairwise sentence ranking task.
\citet{du-etal-2021-self} present a data augmentation method where given a small, labeled training dataset, they retrieve additional training samples from a large unlabeled dataset and then label these samples automatically with a model trained on the original, smaller training dataset. 
The resulting augmented, synthetic dataset can then be used to train another model that generalizes better to unseen data.
Of the related work presented, this approach, also referred to as self-training, is the most similar to the approach we present in this paper. 
The main difference is that \citet{du-etal-2021-self} tailor their approach mainly to domain-specific pre-training, whereas we focus on task-specific fine-tuning.
\citet{xie-etal-2019-selftraining} extend the self-training method by intentionally adding noise to the training process to foster better generalization of the trained models.
Further, they repeat the self-training process several times, so that the model trained on pseudo-labels is again used to create another set of pseudo-labels, which are in turn used to train another model and so on.
As this iterative approach is very resource-intensive in terms of training time, we limit our approach to only one iteration.
However, no inherent limitation prevents our approach from more training iterations.
To the best of our knowledge, there are no published approaches that use neural language models for the particular task of complexity prediction of German texts, but only of English texts~\cite{martinc-etal-2021-supervised}.


\section{A Semi-Supervised Learning Approach for Text Complexity Prediction}
\label{sec:approach}
\begin{figure*}
\centering
\includegraphics[width=1.0\textwidth]{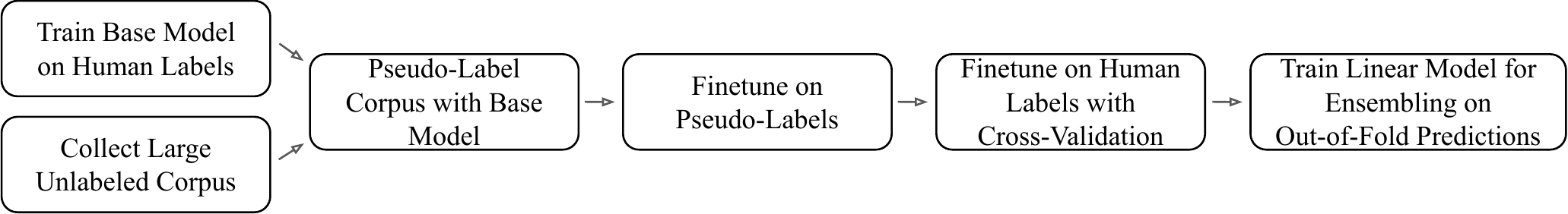}
\caption{Overview of the different steps that comprise the pseudo-labeling, fine-tuning, and ensembling approach.}
\label{fig:overview}
\end{figure*}

Our semi-supervised learning approach uses neural language
models based on the Transformer architecture~\cite{vaswani-etal-2017-attention}. 
As pre-trained models that are not fine-tuned to a specific natural language processing task yet, we use GBERT and GELECTRA models by \citet{chan-etal-2020-germans} and an XLM-RoBERTa model by \citet{conneau-etal-2020-unsupervised}.
Given that the training dataset provided by the shared task organizers is relatively small for fine-tuning these pre-trained models, the core idea of our approach is to increase the number of training samples by automatically generating pseudo-labels. 
Figure~\ref{fig:overview} visualizes the different steps of the entire approach with its three main steps: pseudo-labeling, fine-tuning, and ensembling.
For the implementation of these steps, we use the two open source frameworks FARM\footnote{\url{https://github.com/deepset-ai/farm}} and Haystack\footnote{\url{https://github.com/deepset-ai/haystack}}.

The first step is to create a large corpus of German sentences with varying text complexity to serve as a source for the pseudo-label sentences. This corpus comprises the following resources:
\begin{itemize}
    \item eight random subparts of a German Wikipedia dump (about 8 percent of all 2.3 million German Wikipedia articles as of 2019),\footnote{The most recent dump is available online: \url{https://dumps.wikimedia.org/dewiki/20220720/}}
    \item 130,000 news articles from the German news platform Zeit Online,\footnote{\url{https://www.zeit.de}}
    \item three million sentences from German newspaper texts as part of the Leipzig Corpus Collection~\cite{goldhahn-etal-2012-building},
    \item the Geo/Geolino/Tagesschau/Logo  corpus~\cite{weiss-meurers-2018-modeling},
    \item the Corpus Simple German (all subsets except for Klexikon),\footnote{\url{https://daniel-jach.github.io/simple-german/simple-german.html}}
    \item the Klexikon~\cite{aumiller-gertz-2022-klexikon}, and
    \item the Hurraki dictionary for plain language.\footnote{\url{https://hurraki.de}}
\end{itemize}
Combining these datasets results in a total of 12,955,913 sentences. As some of them appear more than once, this corresponds to 12,562,164 distinct sentences. 
Each of them is embedded using a SentenceTransformers \texttt{msmarco-distilbert} model\footnote{\texttt{sentence-transformers/msmarco\-distilbert\-multilingual-en-de-v2-tmp\-lng-aligned}}~\cite{reimers-gurevych-2019-sentence} and added to an OpenSearch index. Further, we fine-tune a \texttt{deepset/gbert-large} model on the task of sentence complexity on all of the provided training labels as a baseline. Subsequently, to get our set of pseudo-labels, we embed each of the sentences in the provided training set with the same SentenceTransformers \texttt{msmarco-distilbert} model and retrieve the 500 most similar sentences from our large corpus of German sentences as potential pseudo-labels. The baseline complexity scorer model produces a complexity score for each potential pseudo-label. To keep roughly the same distribution as in the original training dataset, we filter the generated pseudo-labels in the following way: we keep only those sentences whose predicted score does not deviate more than the standard deviation of the ratings of the original sentence used to retrieve the 500 potential pseudo-labels. This filtering results in a total of 228,796 pseudo-labels. Table~\ref{table:Datasets} lists the number of pseudo-labels originating from the different data sources.

\begin{table*}[t]
\centering
\caption{Number of pseudo-labeled sentences with the average length in characters, and the average mean opinion score per data source. The text complexity of the sources differs with Wikipedia and Hurraki being the most, respectively least difficult.}
\begin{tabular}{lrrr}
        \toprule
    \textbf{Data Source} & \textbf{\#Sentences} & $\pmb{\varnothing}$\textbf{Length} & $\pmb{\varnothing}$\textbf{MOS}\\
    \midrule
     \textsc{German Wikipedia} & 137,228 & 133 & 3.0\\
     \textsc{Zeit Online} & 47,613 & 108 & 2.5\\
     \textsc{3 Million News Sentences} & 25,928 & 110 & 2.6\\
     \textsc{GEO/GEOlino/Tagesschau/Logo} & 7,971 & 98 & 2.5\\
     \textsc{Corpus Simple German} & 4,896 & 93 & 2.3\\
     \textsc{Klexikon} & 3,600 & 75 & 2.0\\
     \textsc{Hurraki} & 1,559 & 43 & 1.5\\
    \bottomrule
\end{tabular}
\label{table:Datasets}
\end{table*}

The pseudo-labels are used to fine-tune different Transformer-based models on the task of complexity scoring. We fine-tune \texttt{deepset/gelectra-large}, \texttt{deepset/gbert-large} and \texttt{xlm-roberta-large} using three different seeds for each model, resulting in a total of nine models. Subsequently, we fine-tune each of these models using five-fold cross-validation with the original training set, resulting in a total of 45 models.
Finally, to combine these 45 models into an ensemble providing a single prediction score per data sample, we train linear regression models on the out-of-fold predictions from the previous cross-validations.


\section{Experiments}
\label{sec:experiments}
We evaluate four different settings using five-fold cross-validation:
\begin{itemize}
    \item a baseline \texttt{deepset/gbert-large} model fine-tuned on the provided training set,
    \item an ensemble of nine models fine-tuned only on pseudo-labels and with three different random seeds, scores aggregated by mean (\texttt{deepset/gbert-large}, \texttt{deepset/gelectra-large}, \texttt{xlm-roberta-large} 
    \item an ensemble of 45 models fine-tuned on pseudo-labels and the provided training set, with scores aggregated by mean, and
    \item an ensemble of 45 models fine-tuned on pseudo-labels and the provided training set, with scores aggregated by a linear model.
\end{itemize}

We submitted the predictions of each of the last three settings to the shared task competition.\footnote{\url{https://codalab.lisn.upsaclay.fr/competitions/4964}}

The first setting serves as our baseline with a \texttt{deepset/gbert-large} model fine-tuned on the provided training data. The model is trained on each of the cross-validation folds using early stopping for a maximum of four epochs. Each training run was tracked using MLflow and can be found \href{https://public-mlflow.deepset.ai/#/compare-runs?runs=[%22189a619c257d42bb81740547d982bc62%22,%22a96714b9e9b64eb79f40dc771fc43dd8%22,%22d86720d85b4845909782756c3b511a77%22,%22bad0a5b2197e4c42ad55e2c379c4cf18%22,\%22b20dac7cb1e942cbb7ccc59357898080%22]&experiment=560}{here}.

The second setting is an ensemble of the language models \texttt{deepset/gbert-large}, \texttt{deepset/gelectra-large} and \texttt{xlm-roberta-large}. Each of these models is fine-tuned for two epochs on the pseudo-labels described in Section~\ref{sec:approach} using three different random seeds. This results in an ensemble of nine models. One training run takes approximately three hours on an NVIDIA Tesla V100 GPU with 16 GB of RAM. 

For the third and fourth setting, we fine-tune the resulting models of the previous step on the provided training dataset. To further increase the number of models in the ensemble, we perform five-fold cross-validation on each of the nine models, resulting in a total of 45 models. Again, each training run was tracked using MLflow and can be found \href{https://public-mlflow.deepset.ai/#/compare-runs?runs=[%22d4d4ba81b7084ee29f51c07248f841a1%22,%22cb51b7a5ae5445f4a908c0229925cb54%22,%2299212c9a951543f8aaea7a2c64fd63ef%22,%220928389a5c914b8581113358a29340c4%22,%2227ac05a160ed428eb0f1b0f07d2a73ed%22,%22f2240f0f8fed4f6590420321a80e52fd%22,%22362f8a5343a14aa2b13d00f513e19052%22,%220539b876f74a4a508100816a476de8e2%22,%22c54a757edc7444b79566b395087aa7b5%22,%2209557849f8b54e1aab247fa00283b78d%22,%223cf22a6a8ba14bf0b64d1ccbaf6bf3f3%22,%223a3d19acaf584d999f5c07820e9a116d%22,%2267890d23fb7349d68ded69901e5b4a4f%22,%22ad96d9e0dc1b403b880d51149c1231fd%22,%228f2dc5c7cc29489d9698d2270297e292%22,%22ea178a505a5b43fb9c1cb84668403d90%22,%22f446a239dd0c40edbeda2521827eaa77%22,%22730c33aa3cfb4ad28a23e0cad47471bb%22,%22771763983a534f2e814ed751b502f84a%22,%22230d9af487b44295bad59f57e80e1f13%22,%2237929400152f4d2084116a87b721c939%22,%22e55f70f3da20484889ab5e8810f76f32%22,%220b0a3213a5c449f58afbd5c98989316d%22,%221fb62d0ff447430c9bda9e5b5700318e%22,%22e5af89a3109f4fc9b32d11ed1fbaa3c0%22,%2243d792fee61c4f4d9e604e5921efb40f%22,%2292dee3ad99714e49b563488f6d198bd3%22,%22b40a28ee733a46cda76ceaca85cfb222%22,%22b410aee81f5b45a3947a7fdcff504e10%22,%225497a0f5d64d412baa020fc925f51695%22,%2269317267890c412eb3e6625c3d9617c6%22,%227d97ace6cd7c4dffba1428d15caa21af%22,%228ee83e5bdec8412ba3d9acf288dd02fb%22,%22cb9e6d9f79a44a05a7d590e4bde129bf%22,%22b0a241c8650140839c5b5eb64a05a50a%22,%2248cfa0218fe9408fb288191eaf99b28c%22,%22847dfa3a3155414e8710e11f553015b2%22,%22ab739830d0b141c896c898147916da1d%22,%227e71372bacc3413786089713eadfcf00%22,%22de5418458c444c3d85bf61c5c2f76196%22,%2203741051db1f48c08377f57750547c50%22,%220d4545ecb2a047abbdab64a7145c0f30%22,%2276788ee2ccb845bf9cc3262ebeddab4d%22,%221a6f42548703452c99701b04d6414e65%22,%22a04c3c0dc45e48e48073d93ff52e4f98%22]&experiment=560}{here}. To ensemble these 45 models, we use two different techniques. The third setting aggregates each individual score into a single score by simply taking the arithmetic mean of all scores. The fourth setting trains a linear ridge regression model on the out-of-fold predictions for each model that we trained, resulting in five linear regression models. Applying these linear regression models decreases the number of scores from 45 to 5. To get a single score out of these five scores, we calculate their arithmetic mean.
Table~\ref{tab:hyperparameters} summarizes the hyperparameters that are used to train the models for the different described settings.

\begin{table*}[th]
    \centering
    \caption{Hyperparameters for fine-tuning the language models on the pseudo-labels and the provided training data.}
    \label{tab:hyperparameters}
    \begin{tabular}{lrr}
        \toprule
         \textbf{Hyperparameter}& \textbf{Fine-Tuning on Pseudo-Labels} & \textbf{Fine-Tuning on Training Set} \\
        \midrule
         Learning rate & 1e-5 & 1e-6\\
         LR schedule & linear & linear\\
         Warm-up steps & 10\% & 10\%\\
         Batch size & \multicolumn{1}{p{50mm}}{\raggedleft 20 for \texttt{xlm-roberta-large}, 32 otherwise} & \multicolumn{1}{p{50mm}}{\raggedleft 20 for \texttt{xlm-roberta-large}, 32 otherwise}\\
         Early stopping & \ding{54} & \ding{52}\\
         (Max.) epochs & 2 & 4\\
         Optimizer & Adam & Adam\\
         Max sequence length & 128 & 128\\
         \bottomrule
    \end{tabular}
\end{table*}

Table~\ref{tab:scores} lists the cross-validation RMSE on the provided training set. As expected, the approach of using pseudo-labels in combination with ensembling outperforms the simple baseline. We observe that fine-tuning the models only on the pseudo-labels already outperforms the baseline that uses only the original training data. Performance improves further if the models that were fine-tuned on the pseudo-labels are additionally fine-tuned on the original training data. Moreover, using a linear model to aggregate the individual scores instead of using the plain average does not further improve the final score. 
The best setting, consisting of an ensemble of 45 Transformer models fine-tuned on both the pseudo-labels and the provided training data, with results aggregated using a linear regression model, yields an RMSE of 0.433.

\begin{table*}
\centering
\caption{Cross-validation RMSE.}\label{tab:scores}
\begin{tabular}{lrrrrrr}
\toprule
\textbf{Model} & \multicolumn{1}{c}{\textbf{1}} & \multicolumn{1}{c}{\textbf{2}} & \multicolumn{1}{c}{\textbf{3}} & \multicolumn{1}{c}{\textbf{4}} & \multicolumn{1}{c}{\textbf{5}} & \multicolumn{1}{c}{$\pmb{\varnothing}$} \\
\midrule
Baseline & 0.512 & 0.460 & 0.440 & 0.398 & 0.488 & 0.460\\
Ensemble pseudo-labels only & 0.500 & 0.462 & 0.381 & 0.450 & 0.442 & 0.447\\
Ensemble simple mean aggregation & 0.491 & 0.443 & 0.374 & 0.443 & 0.426 & 0.435 \\
Ensemble linear model aggregation & 0.445 & 0.455 & 0.405 & 0.443 &  0.418 & \bftab 0.433 \\
\bottomrule
\end{tabular}
\end{table*}

\section{Conclusion}
\label{sec:conclusion}
In this paper, we presented our submission to the Text Complexity DE Challenge 2022.
We leveraged pseudo-labeled sentences from Wikipedia and several other publicly available, unlabeled corpora.
Based on the labeled training dataset from the shared task and the additional pseudo-labeled data, we fine-tuned Transformer-based neural language models.
Our best ensemble model achieved an RMSE of 0.433 in cross-validation on the public dataset without third-order mapping and an RMSE of 0.454 on the private test dataset with third-order mapping (0.484 without third-order mapping).
For future work, our trained model could be used to create more pseudo-labels for another iteration of the entire approach, presumably resulting in a model that generalizes even better to unseen test data.

\bibliography{anthology,acl2020}
\bibliographystyle{acl_natbib}
\end{document}